\documentclass[pdflatex,sn-mathphys]{sn_jnl}

\jyear{2021}%

\theoremstyle{thmstyleone}%
\theoremstyle{thmstyletwo}%

\theoremstyle{thmstylethree}%
\newcommand{\ourmethod}{mERE}

\begin{document}

\title[Article Title]{Multilingual Entity and Relation Extraction from Unified to Language-specific Training}


\author[1]{\fnm{Zixiang} \sur{Wang}}\email{wangzixiang@buaa.edu.cn}

\author[1]{\fnm{Jian} \sur{Yang}}\email{jiaya@buaa.edu.cn}

\author[1]{\fnm{Tongliang} \sur{Li}}\email{tonyliangli@buaa.edu.cn}

\author[1]{\fnm{Jiaheng Liu}}\email{liujiaheng@buaa.edu.cn}

\author[1]{\fnm{Ying} \sur{Mo}}\email{moying@buaa.edu.cn}

\author[1]{\fnm{Jiaqi} \sur{Bai}}\email{bjq@buaa.edu.cn}

\author[2]{\fnm{Longtao} \sur{He}}\email{hlt@cert.org.cn}

\author*[1]{\fnm{Zhoujun} \sur{Li}}\email{lizj@buaa.edu.cn}

\affil[1]{\orgdiv{State Key Lab of Software Development Environment}, \orgname{Beihang University}, \city{Beijing}, \state{Beijing}, \country{China}}

\affil[2]{\orgdiv{National Computer Network Emergency Response Technical Team/Coordination Center of China}, \city{Beijing}, \state{Beijing}, \country{China}}


\abstract{Entity and relation extraction is a key task in information extraction, where the output can be used for downstream NLP tasks. Existing approaches for entity and relation extraction tasks mainly focus on the English corpora and ignore other languages.
Thus, it is critical to improving performance in a multilingual setting. 
Meanwhile, multilingual training is usually used to boost cross-lingual performance by transferring knowledge from languages (e.g., high-resource) to other (e.g., low-resource) languages.
However, language interference usually exists in multilingual  tasks as the model parameters are shared among all languages. In this paper, we propose a two-stage multilingual training method and a joint model called Multilingual Entity and Relation Extraction framework (\ourmethod{}) to mitigate language interference across languages. Specifically, we randomly concatenate sentences in different languages to train a Language-universal Aggregator (LA),
which narrows the distance of embedding representations by obtaining the unified language representation.
Then, we separate parameters to mitigate interference via tuning a Language-specific Switcher (LS),
which includes several independent sub-modules to refine the language-specific feature representation.
After that,
to enhance the relational triple extraction, the sentence representations concatenated with the relation feature are used  to recognize the entities. 
Extensive experimental results show that our method outperforms both the monolingual and multilingual baseline methods. Besides, we also perform detailed analysis to show that \ourmethod{} is lightweight but effective on relational triple extraction and \ourmethod{} is easy to transfer to other backbone models of multi-field tasks,
which further demonstrates the effectiveness of our method.
}

\keywords{Joint extraction, Information extraction, Multilingual entity and relation extraction, Relational triple}

\maketitle

\section{Introduction}\label{sec1}
Entity and relation extraction (ERE) contains two sub-tasks called named entity recognition (NER) \cite{DBLP:journals/corr/HuangXY15,DBLP:conf/acl/WangSCC20,DBLP:journals/tacl/JoshiCLWZL20,DBLP:conf/ijcai/Tan0Z0Z21} and relation classification (RC) \cite{DBLP:conf/conll/LevySCZ17,DBLP:conf/coling/LiWZZYC20},  which is the fundamental step of automatic knowledge graphs (KGs) \cite{DBLP:journals/inffus/LinMLXC23} construction, knowledge discovery and intelligent question answering system. The results of ERE are typically described as a relational triple $(h,r,t)$, where $h$ and $t$ are the head entity and the tail entity, respectively, and $r$ denotes the relation between them. 
For example, 
for the sentence ``Big Ben is in UK.'' with a predefined relation called ``Locate\_in'',
an ideal relational triple of this sentence is expressed as (Big Ben, Locate\_in, UK).

\begin{figure}[h]
\centering
\includegraphics[width=0.3\textwidth]{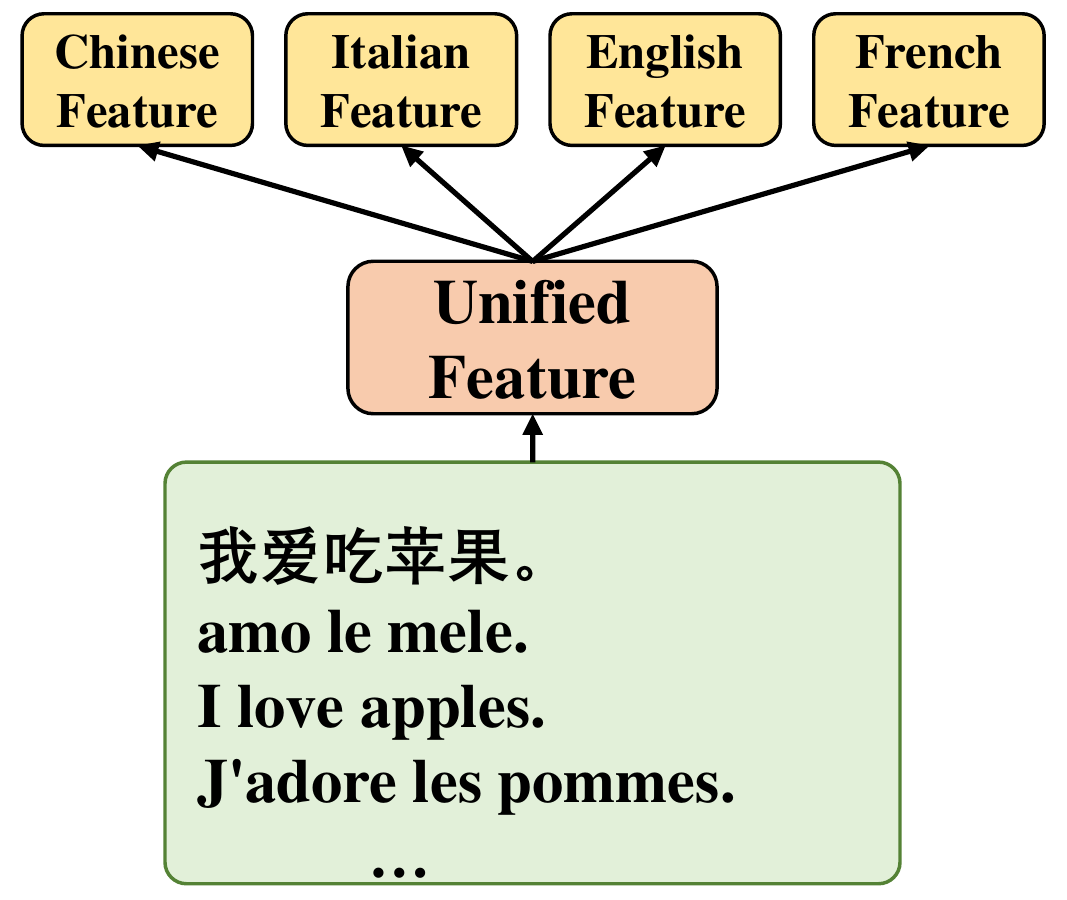}
\caption{This example includes 4 sentences from different languages,
which express the same meaning. The four arrows represent four independent sentence representations extracted from different languages.} \label{fig1}
\end{figure}

As a large amount of data is available from different languages on the Internet,
it is important to utilize such valuable resources and develop multilingual entity and relation extraction models, which can operate across language barriers. 
However, most existing methods propose to solve ERE on English corpora, which can only deal with the monolingual extraction task.
The main reason is that many languages suffer from the scarcity of corpora in ERE. Thus, multilingual training is proposed to help each other in a shared model, where the well-trained knowledge of high-resource languages can be transferred to low-resource languages with a small amount of data. 
Recently, \cite{DBLP:conf/eacl/SegantiFSSA21} propose a multilingual dataset called SMiLER, which is the first work to apply both monolingual and multilingual training. 
The authors in \cite{DBLP:conf/eacl/SegantiFSSA21} introduce the multilingual entity and relation extraction model (i.e., HERBERTa) without considering interference across languages. 
However, such language interference is prevalent in multilingual tasks because of parameter sharing \cite{DBLP:conf/emnlp/WangLT20,DBLP:journals/corr/abs-2104-07358,HLT_MT}.
As shown in Figure~\ref{fig1}, to mitigate interference among languages, we propose to extract the feature representation of the corresponding language sentence. First, to facilitate the cross-lingual transfer among different languages, multilingual representations are supposed to be closed under similar semantics using cross-lingual sentence-level concatenation. Then, based on the shared multilingual parameters, the language-specific representations derived from the independent modules can mitigate interference among multiple languages. 
Specifically, we propose a two-stage multilingual training method and an effective model called multilingual Entity and Relation Extraction framework (\ourmethod{}) to address the multilingual ERE task. In the first stage, we utilize a cross-lingual encoder to encode different language sentences and extract relations directly. Then, we train the joint model with our Language-universal Aggregator (LA) to generate the unified language feature,
which narrows the distance of similar semantic representation across languages. 
LA consists of a self-attention layer and is trained by random multi-sentences concatenation,
which is used to learn semantic similarities in multilingual training. In the second stage, to alleviate the interference among languages, we freeze the parameters of LA and cross-lingual encoder in the first stage and optimize the independent parameters via fine-tuning the model with a Language-specific Switcher (LS), which consists of several independent sub-modules to produce the specific language features. 
Meanwhile, a selection mechanism is applied to choose the optimal group of sub-modules from LS, which enables the sub-module to share the same parameters with a certain group of languages. Such an automatic sub-module selection mechanism saves many model parameters when the number of languages is large. After that, each token representation is concatenated with the relation representation to enhance the recognition of the positions of entities in a sentence. Finally, 
in \ourmethod{}, we adopt joint training to mitigate the error propagation problem. 

We conduct extensive experiments on the SMiLER benchmark of 14 languages with 36 relations (including no\_relation) in total. The experimental results demonstrate that our method outperforms previous monolingual and multilingual ERE baseline methods by a large margin across languages,
which demonstrates that our method can effectively mitigate language interference by improving representation quality among languages. 
Besides, we conduct detailed experiments to analyze how our method affects relational triple extraction. 
Moreover, our method is simple but effective,
and it is also easy to transfer to different backbone models of multi-field tasks with lightweight modules.

\section{Related Work}\label{sec2}

\textbf{Information Extraction} 
Information extraction mainly focuses on extracting knowledge from unstructured text. A well-known system called Never-Ending Language Learner was reading the Web for almost 10 years to collect new instances of pre-deﬁned relations and entity types \cite{DBLP:journals/cacm/MitchellCHTYBCM18}. Instead of the pre-defined entity and relation types, Open Information Extraction (OpenIE) has also attracted much attention during the past decade. A notable example is TextRunner \cite{DBLP:conf/naacl/YatesBBCES07}, which utilizes a syntactic parser to extract triples from the Internet automatically. Many systems have been proposed subsequently, such as rule-based systems \cite{DBLP:conf/emnlp/FaderSE11,DBLP:conf/emnlp/MesquitaSB13,DBLP:conf/emnlp/WhiteRSVZRRD16} and clause based systems \cite{DBLP:conf/www/CorroG13,DBLP:conf/acl/AngeliPM15}. Recent supervised methods are divided into three categories based on different architectures: (1) Generation-based models are typically sequence-to-sequence structure \cite{DBLP:conf/acl/CuiWZ18,DBLP:conf/acl/KolluruARMC20,DBLP:conf/aaai/Li}. (2) Sequence labeling-based models using Begin Inside Outside (BIO) or Subject Relation Object None (SRON) to label every word in a sentence \cite{DBLP:conf/naacl/StanovskyMZD18,DBLP:conf/emnlp/KolluruAAMC20}. (3) Span-based model takes advantage of span level feature which can be sufficiently exploited \cite{DBLP:conf/aaai/ZhanZ20}.


\noindent\textbf{Entity and Relation Extraction}
Early entity and relation extraction tasks use a pipeline approach, which are two separate subtasks including named entity recognition and relation classification. \cite{DBLP:conf/emnlp/SocherHMN12} first works on Recurrent Neural Network (RNN) based model for extraction, capturing the semantics of the entity and its adjacent phrases through parsing trees. While \cite{DBLP:conf/emnlp/HashimotoMTC13} uses a syntactic tree-based RNN model to add weights to the important phrases. \cite{DBLP:conf/coling/ZengLLZZ14} first used a Convolutional Neural Network (CNN) structure to fuse the extracted word and sentence level features for extraction work. \cite{DBLP:conf/emnlp/XuFHZ15} uses a CNN structure based on a dependency tree to improve the performance. However, the pipeline approach has inevitable deficiencies: (1) The architecture ignores the interactions between entities and relations, causing the error propagation problem. (2) Some of the extracted entities are redundant in the named entity recognition phase, resulting in a degradation of performance in the relation classification phase.

Most studies focus on the joint approach, which models entity recognition and relation classification in the same network and naturally relieves error propagation problem. The initial joint models are feature-based methods that heavily rely on NLP tools and manual efforts \cite{DBLP:conf/conll/RothY04,DBLP:conf/conll/KateM10,DBLP:conf/coling/YuL10,DBLP:conf/acl/LiJ14}. Recent joint models are typically neural network-based methods, which benefit from their excellent feature learning capability. SPTree \cite{DBLP:conf/acl/MiwaB16} is the first joint model based on the neural network method. Due to the two subtasks decoding with independent decoders but sharing parameters of the same encoding layers, this architecture also is known as parameters sharing. Following such kind of structure, \cite{DBLP:journals/ijon/ZhengHLBXHX17} proposed an LSTM-based network that decodes entities and a CNN network to classify relations. \cite{DBLP:conf/aaai/TanZWX19,DBLP:conf/ijcai/LiuCWZLX20} employ CRF to improve performance of entity recognition. \cite{DBLP:conf/emnlp/WaddenWLH19,DBLP:conf/ecai/EbertsU20,DBLP:conf/coling/JiYLMWTL20,DBLP:journals/nca/QiaoZHFZC22} use a pre-trained model called bidirectional encoder representation from transformers (BERT) to improve the accuracy of entity recognition. \cite{DBLP:journals/apin/LaiCWYZ22} proposes a multi-feature fusion sentence representation and decoder sequence annotation to handle the overlapping triples which are overlapped with one or two entities. Another architecture is joint decoding, which extracts entity pairs and corresponding relations simultaneously in one stage. NovelTagging \cite{DBLP:conf/acl/ZhengWBHZX17} first proposes a tagging scheme to implement a joint decoding manner. But it cannot figure out the overlapping problem. The sequence-to-sequence scheme \cite{DBLP:conf/acl/LiuZZHZ18,DBLP:conf/emnlp/ZengHZLLZ19,DBLP:conf/aaai/NayakN20,electronics11101535} models relational triples as a sequence, which can naturally deal with the nested entity and overlapping problem.

\noindent\textbf{Multilingual Models} 
Multilingual models are a type of model that performs cross-lingual transfer among different languages, such as multilingual pre-training \cite{DBLP:conf/naacl/DevlinCLT19,ALM,conneau-etal-2020-unsupervised,GanLM,liu2022cross} and machine translation \cite{UM4,HLT_MT,wmt2021_microsoft,guo2022lvp}. Specifically, mBERT pre-trained on 104 languages in Wikipedia has a strong ability for cross-lingual transfer. Multilingual neural machine translation (MNMT) trains a single NMT model in multiple language pairs supporting translation directions between multiple languages by sharing parameters \cite{DBLP:conf/naacl/FiratCB16,DBLP:journals/tacl/JohnsonSLKWCTVW17,DBLP:journals/jmlr/FanBSMEGBCWCGBL21,DBLP:conf/emnlp/LinPWQFZL20}. Early studies mainly utilize high-resource languages to help low-resource languages and even perform zero-shot transfer translation \cite{DBLP:conf/naacl/AharoniJF19,DBLP:conf/acl/ZhangWTS20}. Recent studies focus on designing language-specific components to mitigate the language interference in shared parameters, especially on high-resource pairs \cite{DBLP:conf/rep4nlp/VazquezRTC19,DBLP:conf/emnlp/PhilipBGB20,HLT_MT}. Our method boosts the sentence representation quality from superior unified representation to further language-specific representation.



\noindent\textbf{Multilingual Entity and Relation Extraction} 
Existing entity and relation extraction datasets are insufficient in diversity and size. English is always used to be training corpora. \cite{DBLP:conf/eacl/SegantiFSSA21} presents a new, large and diversiﬁed dataset Samsung MultiLingual Entity and Relation Extraction (SMiLER) dataset to entity and relation extraction both for English and multilingual setting. This is currently the most comprehensive and largest multilingual dataset. 

In this paper, we propose a multilingual entity and relation extraction framework called \ourmethod{} with two-stage training strategies. In the first stage, we concatenate random sentences and use the self-attention mechanism \cite{NIPS2017_3f5ee243} to learn the unified representation across languages. Inspired by MoE \cite{shazeer2017}, we use several sub-modules with a selection mechanism to learn the specific representation of each language in the second stage. Such two-stage learning greatly improves the performance of relational triple extraction.

\begin{figure}[h]
\centering
\includegraphics[width=1.0\textwidth]{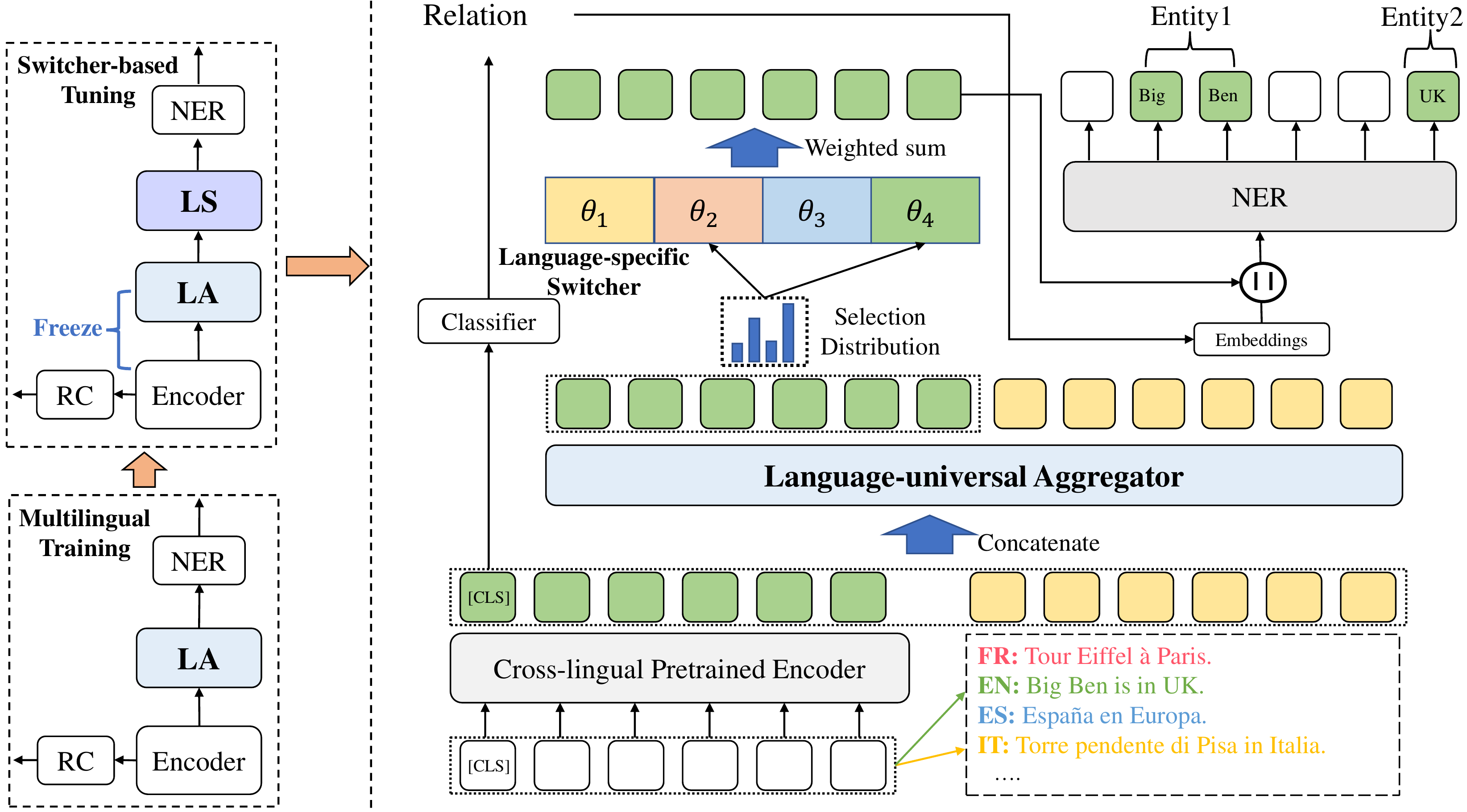}
\caption{The left part shows the two-stage training strategy. The right part is our framework with Language-universal Aggregator (LA) for unified representation generation and Language-specific Switcher (LS) for language-specific feature extraction. We first train the LA with a concatenation of 2 random sentence representations, which are denoted as the green boxes (English) and yellow boxes (Italian) below the figure. Note that each sentence representation is directly regarded as input of LA during the evaluation stage. Then, we freeze part of the parameters and fine-tune the LS with all sub-modules during the training stage. The figure illustrates 4 sub-modules of LS with a top-$2$ strategy during evaluation.} \label{fig2}
\end{figure}

\section{Methodology}\label{sec3}
In this section, we introduce the details of our training method for the multilingual joint extraction model as shown in Figure~\ref{fig2}. We propose a two-stage training strategy. In the first stage, we train a Language-universal Aggregator (LA) for learning the unified representations among multiple languages. In the second stage, we freeze the parameters and fine-tune the Language-specific Switcher (LS), which is applied to select specific feature representations of various languages. 

\subsection{Task Formulation}
The goal of multilingual joint entity and relation extraction aims to identify all possible relational triples from sentences in different languages. Formally, given a sentence $X$ from multilingual corpora $D=\{{D_n}\}_{n=1}^N$, where $N$ represents the number of the all languages $L_{all}=\{{L_n}\}_{n=1}^N$. The probability of the target triple $Y=\{s,r,o\}$ is defined as below:
\begin{equation}
P(Y\mid X) = p(r\mid X;\phi)p(s,o\mid X,r;\varphi),\label{eq1}
\end{equation}
where $r$ denotes relation, $s$ and $o$ are subject (head entity) and object (tail entity), respectively. $p(r\mid X;\phi)$ means relation is only related to sentence $X$, and $p(s,o\mid X,r;\varphi)$ means the entity pair $(s,o)$ is related to both sentence $X$ and the relation $r$ that they shared. 

\subsection{Language-aggregation Training}\label{subsec3}
We train the model with Language-universal Aggregator (LA) to learn the unified representation, which effectively narrows the distance of semantic representations across different languages. To obtain context representations of each token from the multilingual sentences, we utilize the cross-lingual pre-trained encoder for building a multilingual model. Given the sentence $X^{L_n} = \{x_1^{L_n},\dots,x_i^{L_n},\dots,x_m^{L_n}\}$ with $m$ tokens (including \texttt{[CLS]}, \texttt{[SEP]} and \texttt{[PAD]}), $x_i^{L_n}\in \mathbb{R}^d$ is the $i$-th token embedding and $d$ is the embedding size. The whole sentence is encoded by the cross-lingual pre-trained encoder:
\begin{equation}
    h^{L_n} = \mathcal{H}(X^{L_n};\phi), \label{eq2}
\end{equation}
where $h^{L_n} = \{h_1^{L_n},\dots,h_i^{L_n},\dots,h_m^{L_n}\}\in\mathbb{R}^{m\times d}$ represents the encoded representation and $d$ is the hidden size. $\mathcal{H}$ denotes the cross-lingual pre-trained encoder. 
 Meanwhile, a relation classifier $W^r\in\mathbb{R}^{d\times U}$ is used to project pooled output vector $h_{p}$ (from the \texttt{[CLS]} token) to the relation $r_{c}$, where $U$ is the number of relation types. The relation extraction is defined as:
 \begin{equation}
     r_c = h_{p}W^r, \label{eqr}
 \end{equation}
 
To better learn the unified semantic representation among multiple languages, we randomly sample $s$ sentences of different languages from the training corpora to generate the cross-lingual representations using Equation~\ref{eq2} and concatenate them to obtain $h_{cat}=[h_1^{L_{X_1}},\dots,h_i^{L_{X_i}},\dots,h_s^{L_{X_s}}]$,
where $L_{X_i}$ denotes the language symbol of the $i$-th sentence. Considering that each token needs to capture the dependency of inner-sentence and acquire semantic similarity representation of inter-sentence among languages, we train LA which applies the self-attention mechanism for fusing the information of the given concatenated representation: 
\begin{equation}
     \hat{h}_{cat}=\mathtt{SF}(\frac{QK^T}{\sqrt \epsilon})V \label{eq3} 
\end{equation}where $Q=h_{cat}W_q$, $K=h_{cat}W_k$ and $V=h_{cat}W_v$. $\mathtt{SF}$ represents the softmax operation. The three-parameter matrices $W_q$, $W_k$, and $W_v$ are trainable. The term $1/\sqrt \epsilon$ is the scaling factor. $\hat{h}_{cat}=\{\hat{h}_1^{L_{X_1}},\dots,\hat{h}_i^{L_{X_i}},\dots,\hat{h}_s^{L_{X_s}}\}$ and $\hat{h}_i^{L_{X_i}}$ is $i$-th element. Instead of using language-specific features generated via Equation \ref{eq7}, we directly utilize each element representation in $\hat{h}_i^{L_{X_i}}$ to train the model via Equation \ref{eq_entity}.


\subsection{Language-specific Training}\label{subsec4}

To acquire features of a specific language, we freeze the parameters of language aggregation and cross-lingual encoder in the first training stage and fine-tune the model with LS. After obtaining the unified representation via LA, we extract the language-specific features via the LS with the selection mechanism from the unified representations. 

Given the language symbol $L_n\in L_{all}(1\leq n\leq N)$ and our LS $\theta=\{{\theta_t\}}_{t=1}^{\mathcal{T}}(1\leq t\leq {\mathcal{T}},1\leq {\mathcal{T}}\leq N)$, our selection mechanism is used to select corresponding sub-modules $\theta_{f(L_n)}$, in which $f(\cdot)$ is a function that maps a language to corresponding LS modules. To design an appropriate map function for our selection mechanism, each sentence is prefixed to the corresponding language symbol, which enables the model to correctly route sentences. Besides, all sub-modules from LS attend to the selection procedure during the training stage, which solves the undifferentiability problem. Specifically, the function $f_{t}(\cdot)$ indicates the probability of selection of sub-module $\theta_t$:
\begin{equation}
f_{t}\left(L_{n}\right)=\frac{\exp \left(e_{t}^{L_{n}}\right)}{\sum_{i=1}^{{\mathcal{T}}} \exp \left(e_{i}^{L_{n}}\right)} \label{eq4} 
\end{equation}where $e_{i}^{L_{n}}$ is $i$-th element of the probability vector $e^{L_n}=E_l[n]W_f$. $E_l\in\mathbb{R}^{N\times d}$ denotes the look-up table for all language prefix embeddings. The router matrix $W_f\in\mathbb{R}^{d\times {\mathcal{T}}}$ is used to project $e^{L_n}$ which are normalized via a softmax distribution over the total ${\mathcal{T}}$ modules. 

For each sub-module $\theta_t$ from $\theta$, we utilize $\mathcal{E}_{\theta_t}(\cdot)$ to transform unified feature representation $\hat{h}^{L_n}$ into language-specific feature branch $\tilde{h}_{\theta_t}^{L_n}$:
\begin{align}
    \tilde{h}_{\theta_t}^{L_n}&=\mathcal{E}_{\theta_t}(\hat{h}^{L_n}) \label{eq_total}
   \\ 
    \mathcal{E}_{\theta_t}(\hat{h}^{L_n})&=\texttt{LN}\left(\sigma(\hat{h}^{L_n}W_u)W_d+\hat{h}^{L_n}\right) \label{eq_details}
\end{align}where $\hat{h}^{L_n}\in\mathbb{R}^{m\times d}$ is an element of $\hat{h}_{cat}$. $W_u\in \mathbb{R}^{d\times b}$ and $W_d\in \mathbb{R}^{b\times d}$ are projection matrices $(b>d)$. $\sigma$ is the ReLU activation function and $\texttt{LN}(\cdot)$ is the layer normalization function. The right part of Figure \ref{fig2} corresponds to Equation \ref{eq_details}.

To ensure gradients are propagated to all sub-modules of LS $\{{\theta_t\}}_{t=1}^{\mathcal{T}}$, we apply the weighted average for obtaining the language-specific feature:
\begin{equation}
\tilde{h}^{L_n}=\sum_{t=1}^{\mathcal{T}} f_{t}(L_n) \mathcal{E}_{\theta_t}\left(\hat{h}^{L_n}\right) \label{eq7}
\end{equation}
Note that for the whole process, function $f_t(L_n)$ in Equation \ref{eq7} permits differentiability of the router. 

In the evaluation stage, it is necessary to prune several sub-module branches with the lowest selection probabilities to obtain the best performance. Therefore, we use the top-$K$ strategy to select the best $k(1\leq k\leq {\mathcal{T}})$ sub-modules with the highest probabilities to generate the language-specific representation. When $k={\mathcal{T}}$ indicates all sub-modules involved in the calculation which means the selection mechanism is the same as the training stage. The mapping process is described as:
$L_n\longrightarrow \{\pi_1^{L_n},\dots,\pi_i^{L_n},\dots,\pi_k^{L_n}\}\in\Pi(k)$, where  $\pi_i^{L_n}$ is one of the sub-module index that corresponds to language $L_n$  and $\Pi(k)$ is the space of all $k$-length combinations of $C_{\mathcal{T}}^k$ in total.

After obtaining the language-specific representation from LS, we create four matrices to recognize the head and tail positions of two named entities. To enhance the accuracy of recognition, we add a relation feature that constrains the extracted entities that are only related to the relevant relation. Formally, given a language-specific representation $\tilde{h}^{L_n}\in\mathbb{R}^{m\times d}$ of the $m$-length sentence and the relation vector $r_e$ retrieved from relation embedding table $E_r\in\mathbb{R}^{I\times d}$, where $I$ is the number of relations, the two entities are recognized as followed:
\begin{align}
entity^{x}&=(\eta ((\tilde{h}^{L_n}\oplus r_e)W_{y}))U_{y} \label{eq_entity}
\end{align}where the symbol collection entity=\{head, tail\}, $x$=\{start,end\} and  $y=\{hs,he,ts,te\}$. We concatenate the relation vector with each token representation to enhance the recognition of entities, namely $\tilde{h}^{L_n}\oplus r_e=\{[\tilde{h}_1^{L_n},r_e],\dots,[\tilde{h}_i^{L_n},r_e],\dots,[\tilde{h}_m^{L_n},r_e]\}\in\mathbb{R}^{m\times 2d}$. 
$W_{y}\in\mathbb{R}^{2d\times d}$ are four down projection matrices and $U_{y}\in\mathbb{R}^{d\times 1}$ are four index projection matrices. $\eta$ denotes $tanh$ activation function. Note that we use ground-truth relation as input in training entity recognition, which conforms to the joint training method in our architecture. 

\subsection{Training Objective}\label{subsec5}
Our model presented in Figure \ref{fig2} is trained jointly on multilingual ERE corpora. We ﬁrst train the model only using a multilingual training strategy for our Language-universal Aggregator. Based on the unified language representation, we fine-tune the model with Language-specific
Switcher for learning the language-specific feature in the next step. The objective is to minimize the two training loss functions which are defined below:
\begin{align}
    \mathcal{L}_{LAT}&=\sum_{m=1}^M \mathbb{E}_{(x, y) \sim D_m}[\mathcal{L}_{ere}(x,y; \Theta)] 
    \\
    \mathcal{L}_{LST}&=\sum_{m=1}^M \mathbb{E}_{(x, y) \sim D_m}[\mathcal{L}_{ere}(x,y; \Theta,\theta)]
\end{align}where $D$ means multilingual entity and relation extraction training corpora and $M$ denotes the number of the samples. $\Theta$ indicates shared parameters and $\theta$ is parameters in LS with selection mechanism. $\mathcal{L}_{ere}$ is the loss function for entity and relation extraction, which is defined as below:
\begin{equation}
    \mathcal{L}_{ere}=\frac{\alpha}{2}(\mathcal{L}^{ \text {start }}_\text {h}+\mathcal{L}^{\text { end }}_\text {h}+\mathcal{L}^{ \text { start }}_\text {t}+\mathcal{L}^{\text { end }}_\text {t})+\beta \mathcal{L}^{\text {rel}}
\end{equation}where each $\mathcal{L}$ with any superscript is a cross-entropy loss. The subscripts with $h$ and $t$ indicate the head entity and tail entity respectively. The $start$ and $end$ of superscripts denote the first token index and last token index of an entity separately. $\mathcal{L}^{rel}$ is the loss function for relation classification. $\alpha$ and $\beta$ are two weights on entity recognition loss and relation classification loss respectively.

\section{Experiments}\label{sec4}
\subsection{Datasets} 
We evaluate our model on the dataset SMiLER \cite{DBLP:conf/eacl/SegantiFSSA21},
which is the largest and most diversified multilingual dataset for multilingual entity and relation extraction tasks with 14 languages from 36 relation types.
The SMiLER consists of about 1.1M annotated sentences from Wikipedia and DBpedia,
which includes English (En), Korean (Ko), Italian (It), French (Fr), German (De), Portuguese (Pt), Nederlands (Nl), Polish (Pl), Spanish (Es), Arabic (Ar), Russian (Ru), Swedish (Sv), Farsi (Fa), Ukrainian (Uk).
The relation types belong to roughly nine domains: location, organization, person, animal, art, device, measurement, event, and no\_relation. The statistics of SMiLER are shown in Table \ref{tab1}. 
As the development set in SMiLER is not publicly available, we only randomly extract the sentences from the training set to create new files with the same split ratio as the original paper.

\begin{table}[h]
\begin{center}
\caption{The statistics of SMiLER dataset. English corpora include full-size, middle-size, and small-size. The languages are ordered from high-resource languages (left) to
low-resource languages (right).}  \label{tab1}%
\resizebox{\linewidth}{!}{ 
\begin{tabular}{@{}lcc|cccccccccccccc@{}}
\toprule
Languages & EN-full & EN-mid  & It & Fr & De & Pt & Nl & En-small & Ko  & Pl & Es & Ar & Ru & Sv & Fa & Uk \\
\midrule
sentences num.    & 748k   & 269k  & 76k & 62k  & 53k  & 45k & 40k  & 35k  & 20k & 17k & 12k & 9k & 7k  & 5k & 3k & 1k \\
relation types    & 36 & 36 & 22 & 22 & 22  & 22 & 22 & 32 & 28 & 22  & 22 & 9 & 8  & 22 & 8 & 7 \\
\botrule
\end{tabular}
}
\end{center}
\end{table}

\subsection{Implementation Details} 
We conduct experiments on SMiLER, 14 languages in total. EN-small is treated as our English corpora. We utilize mBERT as our cross-lingual encoder. We train our model with AdamW, the learning rate is $3$e-$5$ and weight decay is $0.1$. The batch size is set to 16 on Tesla V100 GPU. The hidden size $d$ is 768 and dimension $b$ of projection matrices $W_u$ and $W_d$ is 1024. The max sequence length is 256 and we concatenate 2 sentences during the first training stage. For the second training stage, we freeze most parameters in the first stage except the relation classifier and 8 matrices used to predict entities from Equation \ref{eq_entity}. The sub-module number $\mathcal{T}$ of LS is set to 6 (2 layers for 3 sub-modules and 1 layer for the other). The epoch is set to 5 at the first stage. The max epoch of the second stage is set to 8 with an early stopping mechanism. The loss weights are set to 2 in named entity recognition and 1 in relation classification. 

In the evaluation stage, we set $k=3$ in the top-$K$ strategy to select the sub-modules in LS. We adopt standard micro-$F_1$ metric to calculate scores on the models. The extracted entity pair is regarded as correct if the predictions of the head entity and tail entity are both the same as the ground truth. A triple is treated as correct if the entity pair and the corresponding relation type are all correct. $no\_relation$ type is included in relation prediction. We also add a mask for the relation that is not absent in a language.

\subsection{Baselines} 
As far as we know, the SMiLER is a new dataset and thus only an existing method for multilingual ERE without publishing source code. The relevant task is cross-lingual relation classification,
which is also few in studies. Therefore, we reproduce the following competitive baselines to compare with our proposed approach for a fair comparison:

\begin{itemize}
    \item \textbf{HEBERTa} \cite{DBLP:conf/eacl/SegantiFSSA21}: A multilingual entity and relation extraction framework called Hybrid Entity and Relation extraction BERT, which achieves the state-of-the-art performance on SMiLER. HERBERTa uses a pipeline training manner that combines two independent BERT models. The ﬁrst sub-model classiﬁes the input sequence as one of 36 pre-defined relations (including no\_relation). 
    The relation generated from the first sub-model is then fed to the second BERT and concatenated with the same input sequence as the input of the second model for entity recognition.
    \item \textbf{mBERT} \cite{koksal-ozgur-2020-relx}: A cross-lingual model first uses the mBERT as a backbone for RC, which is trained on 104 languages with the corresponding Wikipedia dumps. We reproduce the results with the code shared at \url{https://github.com/boun-tabi/RELX}
    \item \textbf{MTMB} \cite{koksal-ozgur-2020-relx}: A multilingual pre-training scheme called Matching the Multilingual Blanks (MTMB). The framework shows several advantages against the mBERT on monolingual tasks and achieves significant improvements in cross-lingual transfer.
    Note that this framework is only designed for RC and not adapted to entity and relation extraction. Therefore, we simply modified the output layer of the baseline to conduct the ERE task.
\end{itemize}

In addition to the above baselines, we also build a simplified multilingual joint entity and relation extraction framework called \ourmethod{}-LS-LA as a basic structure which is concatenated relation representation with the sentence representation to enhance the extraction performance. 
\begin{table}[h]
\begin{center}
\caption{The $F_1$ scores of different models. * denotes the model is reproduced by us on our experiment settings. - denotes that the language data is not involved both in the training and the evaluation stage. MONO, EURO, and SVO mean training data in 3 different language groups. The languages are ordered from high-resource languages (left) to low-resource languages (right). The bold font number is the best score in each language.}  \label{tab2}%
\resizebox{\linewidth}{!}{ 
\begin{tabular}{@{}l|c|cccccccccccccc@{}}
\toprule
Test Sets & AVG  & It & Fr & De & Pt & Nl & En & Ko & Pl & Es & Ar & Ru & Sv & Fa & Uk \\
\midrule
HERBERTa*   & 75.5 & \textbf{83.9} & 68.7 & 71.5 & 72.1 & 78.5 & 60.9 & 80.4 & 83.1  & 60.0 & 88.4 & 79.4 & \textbf{84.8} & \textbf{79.6} & 65.0 \\ 
mBERT*\footnotemark[1] & 75.2 & 81.5 & 68.2 & 70.7 & 71.0 & 77.6 & 59.9 & 78.5 & 81.1 & 61.3 & 89.5 & 81.7 & 81.5 & \textbf{79.6} & 70.0 \\
MTMB*\footnotemark[1] & 75.6 & 80.9 & 67.8 & 70.9 & 70.3 & 79.1 & 58.3 & 79.3 & 82.2 & 58.2 & 91.1 & 74.1 & 83.7 & 77.8 & \textbf{85.0} \\ \midrule
\ourmethod{}   & \textbf{77.9} & 81.7 & \textbf{70.3} & \textbf{73.4}  & \textbf{74.3} & \textbf{81.1} & \textbf{62.3} & \textbf{82.7} & 81.6  & 64.7 & \textbf{91.6} & 83.1  & 83.7 & \textbf{79.6} & 80.0 \\
\ourmethod{} (EURO)   & 70.9 & 81.4 & 70.2 & 72.1  & 74.2 & - & 62.2 & - & -  & \textbf{65.2} & - & -  & - & - & - \\
\ourmethod{} (SVO)  & 75.7 & 81.3 & 70.0 & 72.9 & 73.3 & 80.6 & 62.1 & - & 81.0  & 64.7 & - & 83.1  & 83.7 & - & 80.0 \\
\ourmethod{}-LS   & 77.2 & 80.9 & 69.7 & 72.0  & 73.5 & 80.4 & 62.2 & 80.4 & 81.6  & 62.1 & \textbf{91.6} & 83.1  & \textbf{84.8} & 77.8 & 80.0 \\
\ourmethod{}-LS-LA (MONO)   & 70.9  & 81.2 & 68.3  & 67.1  & 68.4 & 77.9 & 58.6  & 79.3 & 79.0 & 48.4 & 90.0 & 72.5  & 80.4 & 66.7 & 55.0 \\
\ourmethod{}-LS-LA   & 76.5 & 81.3 & 69.0 & 71.9  & 71.4 & 80.3 & 60.3 & 76.4 & \textbf{84.2}  & 60.7 & 90.0 & \textbf{83.9}  & 83.7 & 77.8 & 80.0 \\

\botrule
\end{tabular}
}\footnotetext[1]{We modified the output layer to implement the entity recognition to accommodate the ERE task. We train the model in the joint training method. }
\end{center}
\end{table}

\subsection{Models and Languages Comparison} 
The results presented from the Tables are rounded to one decimal place. From Table \ref{tab2}, our method improves multilingual baselines by a large margin over previous baselines. There is a $2.3\%$ improvement on averaged $F_1$ score compared with the previous strongest baseline MTMB which outperforms HERBERTa due to its strong multilingual pre-training scheme. Our \ourmethod{} achieves the best scores on 8 out of 14 languages, especially on high-resource languages. The other 5 out of 6 languages achieve the second-best scores. Surprisingly, even our baseline \ourmethod{}-LS-LA has $0.9\%$ improvement over the MTMB. It seems that our basic structure is more effective on multilingual entity and relation extraction tasks. Compared with \ourmethod{}-LS that only uses LA, our full model \ourmethod{} has nearly $0.7\%$ $F_1$ value improvement on average and yields similar or higher results on 13 languages except for Sv. The improvement can be attributed to our switcher-based language-specific training strategy, which finally extracts accurate information for entity recognition in each language. Compared with our baseline \ourmethod{}-LS-LA, our full model \ourmethod{} has nearly $1.4\%$ $F_1$ value improvement on average which means \ourmethod{}-LS also has nearly $0.7\%$ $F_1$ value improvement on average. All such impressive results demonstrate that our full model \ourmethod{} truly enhances the representation quality and mitigates language interference to a certain extent.

We set several language groups to analyze the impact of different languages: (1)MONO: 14 languages in monolingual training. (2)EURO: It, Fr, Pt, De, Es, En. (3)SVO\footnotemark[2]: EURO, Ru, Sv, Nl, Pl, Uk.\footnotetext[2]{SVO stands for the relative position of the Subject, Verb, and Object in the typical afﬁrmative sentence. We treat Korean, Farsi, and Arabic as non-SVO languages. Arabic is VSO, while Korean and Farsi are SOV.} The default is all languages in multilingual training from Table \ref{tab2}. Compared with \ourmethod{}-LS-LA training in multilingual corpora, we can observe that multilingual training achieves much higher results than \ourmethod{}-LS-LA (MONO) monolingual training from Table \ref{tab2}, especially on low-resource languages. Such as improvements of Uk ($25\%$), Fa ($11.1\%$), and Ru ($11.4\%$). It demonstrates that languages with less training data can benefit most from high-resource languages in multilingual training including ERE tasks. The results of the EURO family group are close to \ourmethod{}. It is worth noting that Es achieves the best score in the EURO group. We conclude that Es benefits a lot from similarities of languages that are in the same language family even with less training data. In the SVO group, we can also visualize that most languages in EURO decrease slightly with the interference of other non-EURO languages. The different language families, or the languages with a big difference in syntactic structures might be the main interference among languages. However, compared with \ourmethod{} (SVO), \ourmethod{} yields the same results on low-resource languages and somewhat higher results on high-resource languages even the three non-SVO (Fa, Ar, and Ko) data involved during the training stage. We suppose that these non-SVO languages which are big different from others and are all low-resource may facilitate distinguishing high-resource languages in learning language-specific features due to each sub-module from LS being independent, without sharing parameters in the same space. Lastly, we also observe some duplicated $F_1$ scores across low-resource languages. This phenomenon is caused by a small number of sentences in test sets. 


\subsection{Entity and Relation Analysis} 
\begin{figure}[h]
\centering
\includegraphics[width=1.0\textwidth]{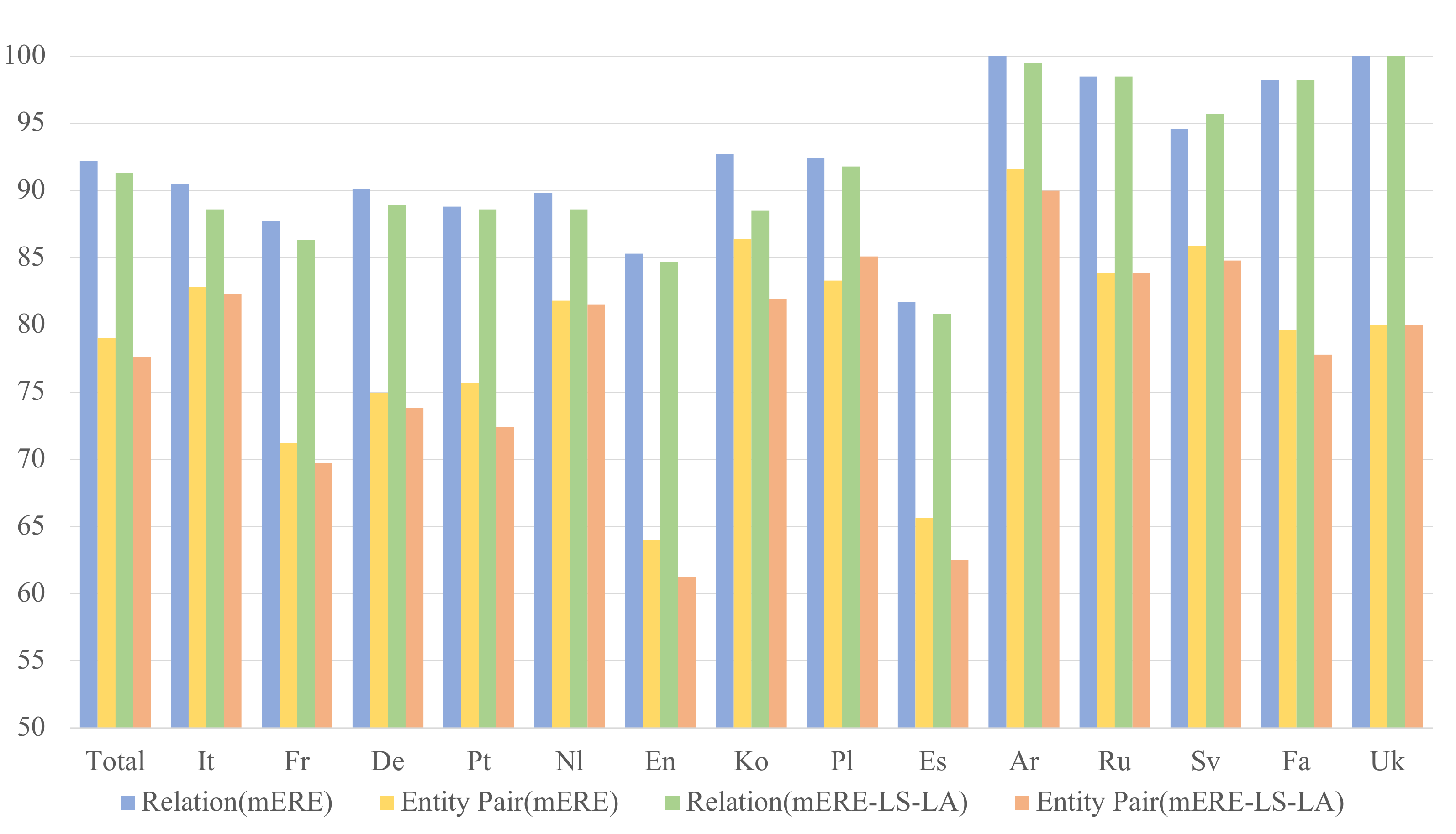}
\caption{The $F_1$ scores of relations and entity pairs on all languages.}\label{fig3}
\end{figure}

Figure \ref{fig3} shows $F_1$ scores of relation and entity pair of \ourmethod{} and \ourmethod{}-LS-LA. We can observe that the relation classification seems to be easier than the named entity recognition. The correctness of entity pair extraction is the main bottleneck of the model performance. With the help of our LA and LS, \ourmethod{} achieves higher results on entity pair recognition compared with \ourmethod{}-LS-LA in general. Surprisingly, we can visualize that the performance of relation classification also has a slight improvement in \ourmethod{}. We conclude that the improvement of the named entity recognition facilitates relation classification. Since information interaction between two sub-tasks can benefit each other in the joint training architecture.

\begin{figure}[h]
\centering
\includegraphics[width=1.0\textwidth]{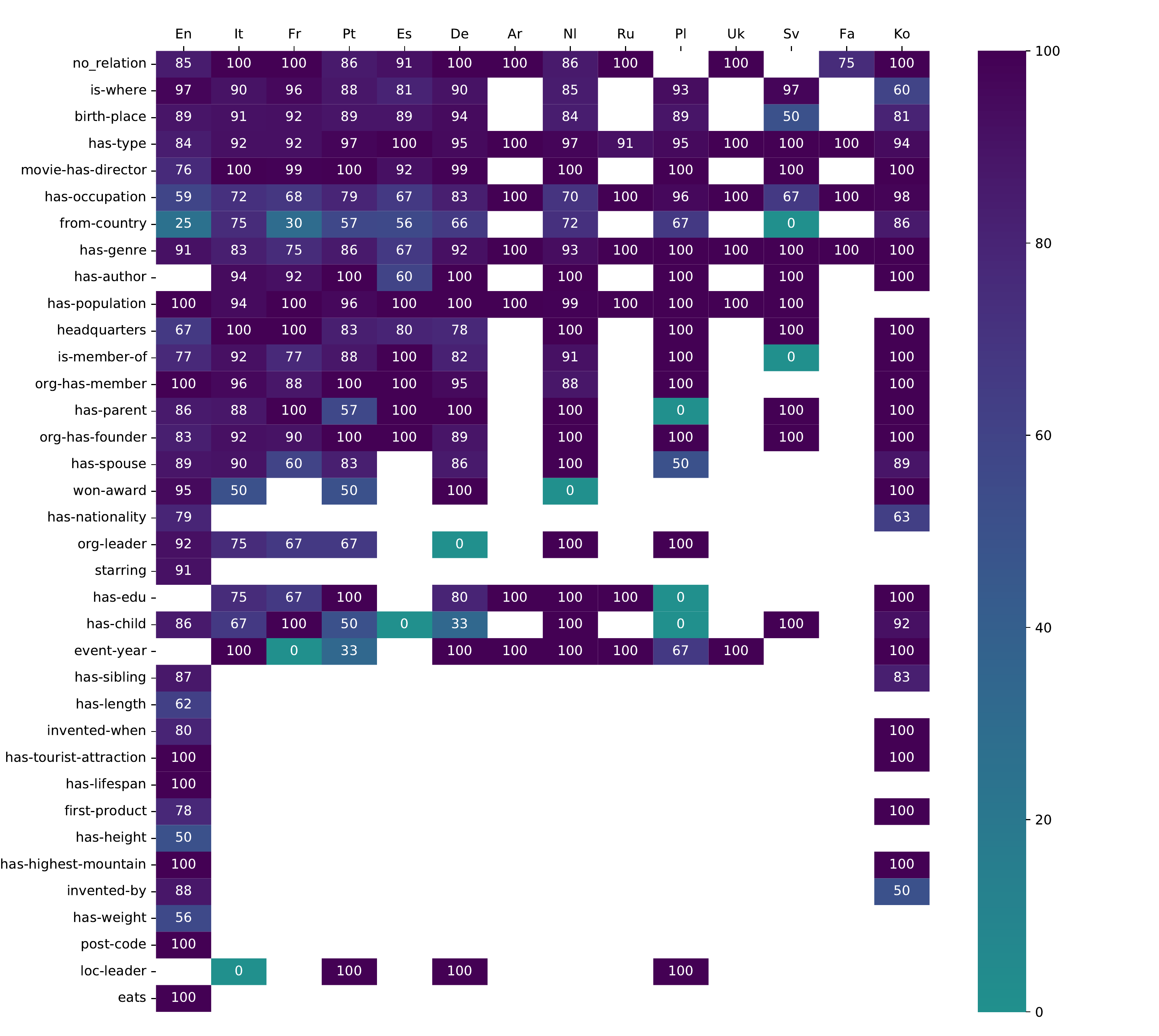}
\caption{The $F_1$ scores of all relation labels on all languages. The darker color means a higher $F_1$ score, while the lighter color means a lower $F_1$ score.}\label{fig4}
\end{figure}

$F_1$ scores of detailed relation labels are shown in Figure \ref{fig4}. Most of the relations achieve higher $F_1$ scores across languages, such as ``no\_relation'' and ``has-type''. Part of relations differs widely across languages, such as relation ``has-child''($F_1=100$ on Nl, $F_1=33$ on De, $F_1=0$ on Es). The big difference is caused by the number of relations of training data in each language. For some relations that occur $F_1=0$ scores, we find out the relations (e.g won-award on Nl. has-parent on Pl. has-child on Es) are only one test sample. Such low results for some languages could be explained by a smaller number of relations in the test set.

\begin{figure}[h]
\centering
\includegraphics[width=0.7\textwidth]{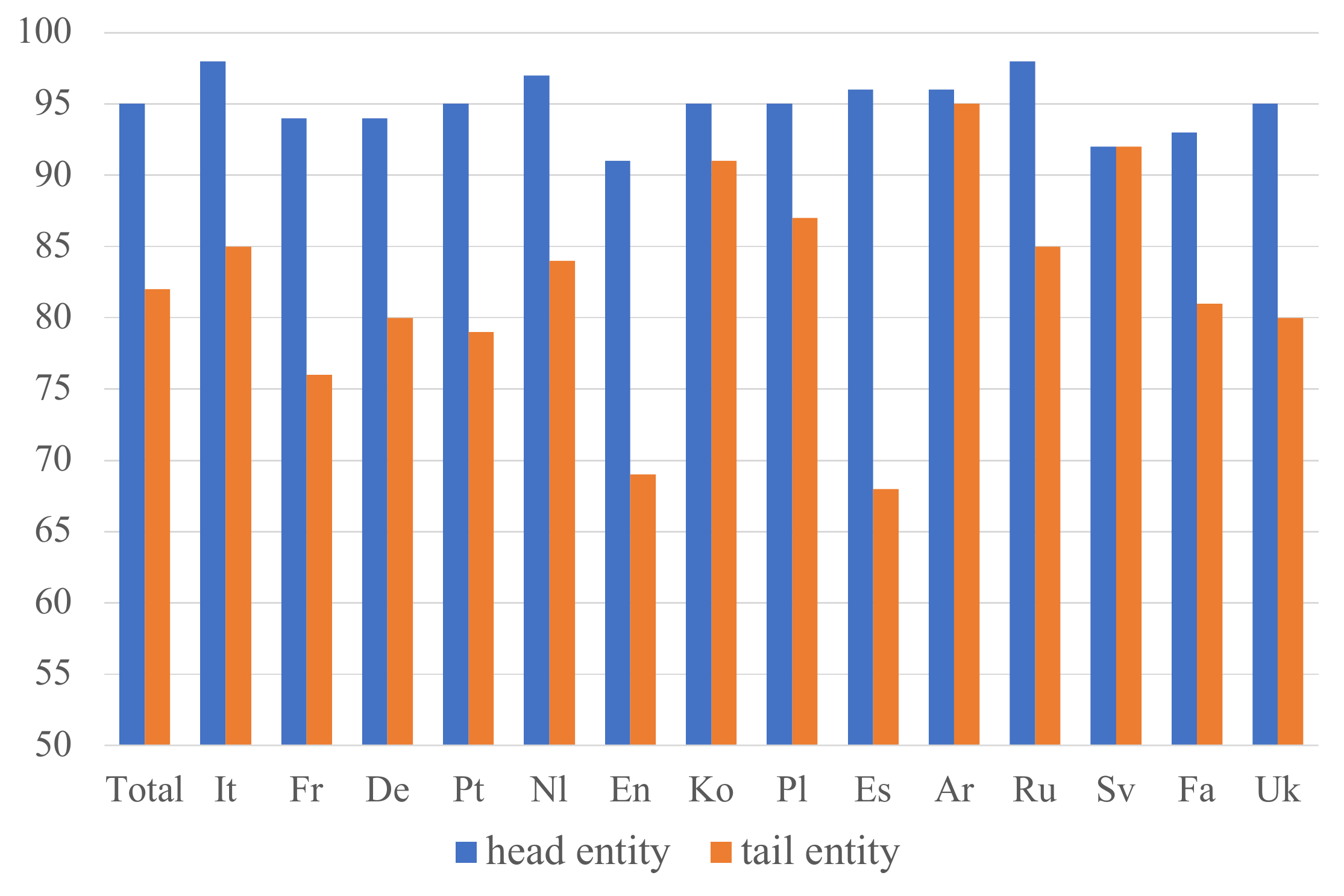}
\caption{The performance of head entities (blue bar) and tail entities (orange bar) on different languages.}\label{fig5}
\end{figure}

Figure \ref{fig5} shows $F_1$ scores of head entities and tail entities. We can observe that $F_1$ scores of head entities are much higher than tail entities among most languages. It seems that head entities are easier to be recognized than tail entities. It is because the head entity always occurs at the beginning position of the sentence and thus the model probably memorizes the position, while the tail entity does not have any consistent position which is hard to predict. 

\subsection{Ablation Study}
\begin{figure}[h]
\centering
\includegraphics[width=0.7\textwidth]{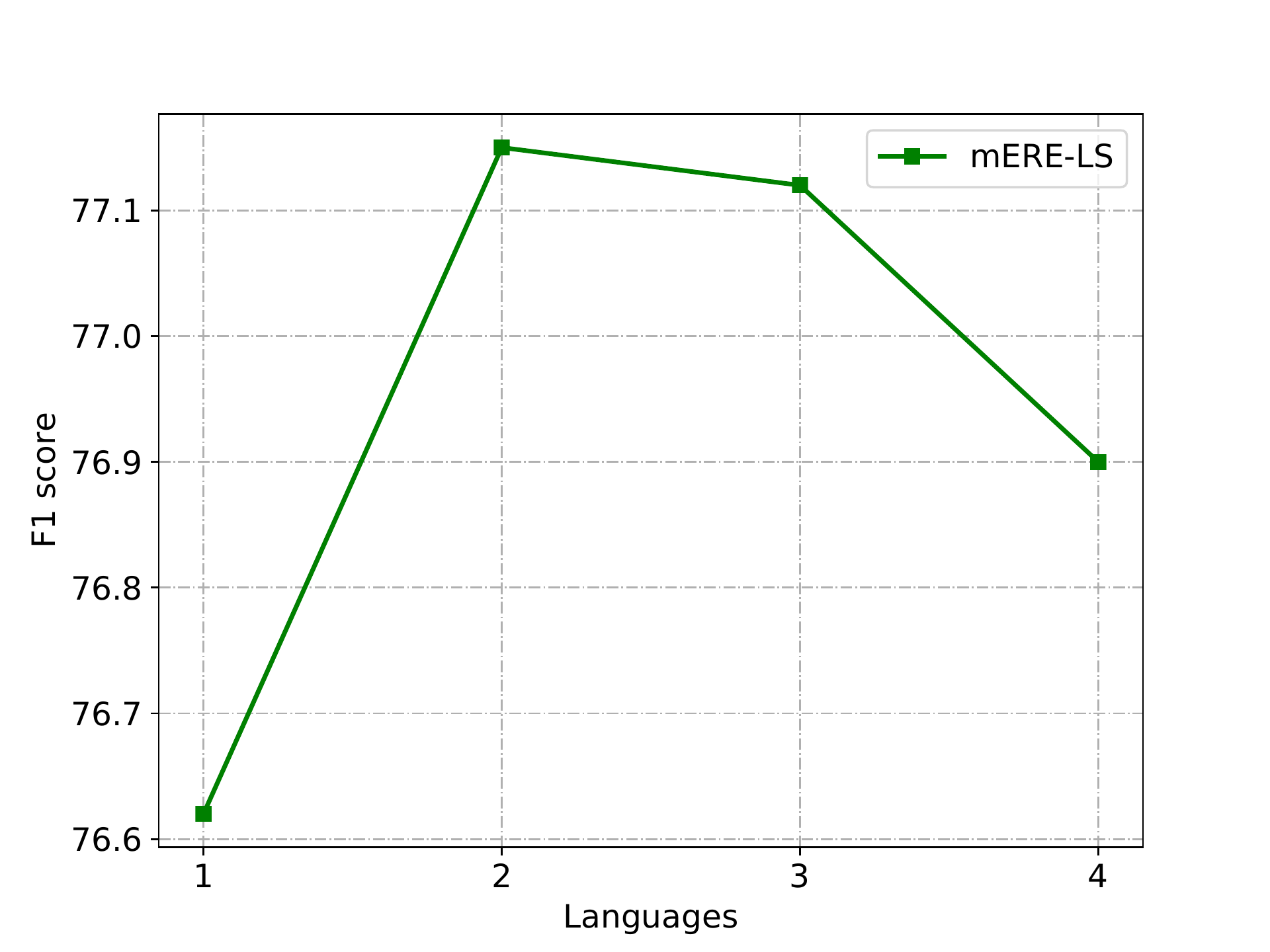}
\caption{The performance of sentences concatenation in the first training stage.}\label{fig6}
\end{figure}

\textbf{Sentences Concatenation} To validate the effect of the number of sentences for learning the unified features among different languages, we conduct several experiments on the different numbers of sentences in concatenation. We learn from Figure \ref{fig6} that there are evident $F_1$ improvements with LA on different concatenation numbers of sentences over only one sentence encoding. The multilingual model obtains the best performance when concatenating with the sentence pair. The increasing number of concatenated sentences has a slight decrease in performance. We conjecture that increasing the number of sentences may also bring somewhat interference. 

\begin{figure}[h]
\centering
\includegraphics[width=0.5\textwidth]{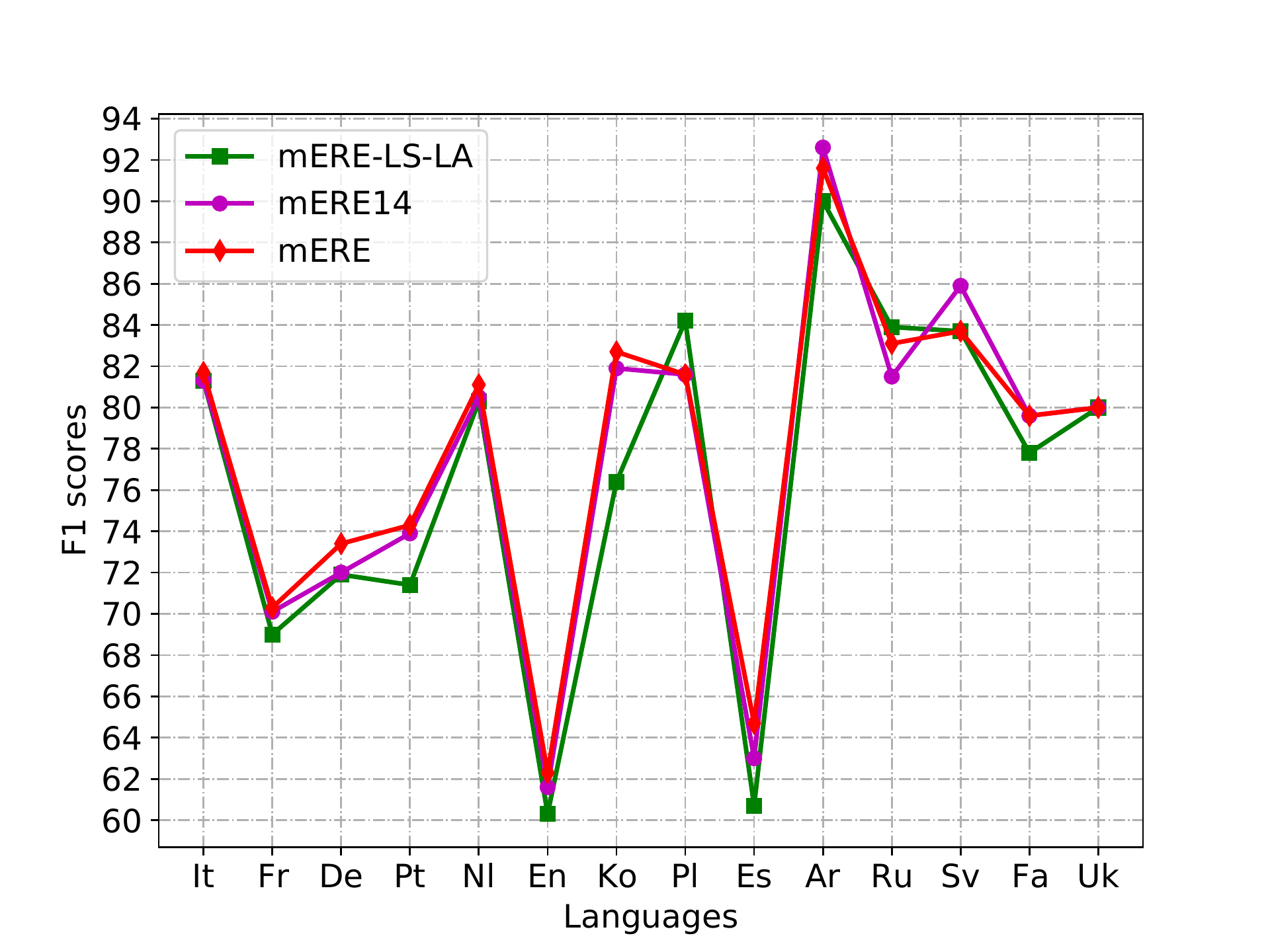}
\caption{The performance of three models on 14 languages. \ourmethod{}14 utilizes 14 one-to-one sub-modules of LS without the selection mechanism. Each sub-module corresponds to a language.}\label{fig7}
\end{figure}

\begin{figure}[h]
\centering
\includegraphics[width=1.0\textwidth]{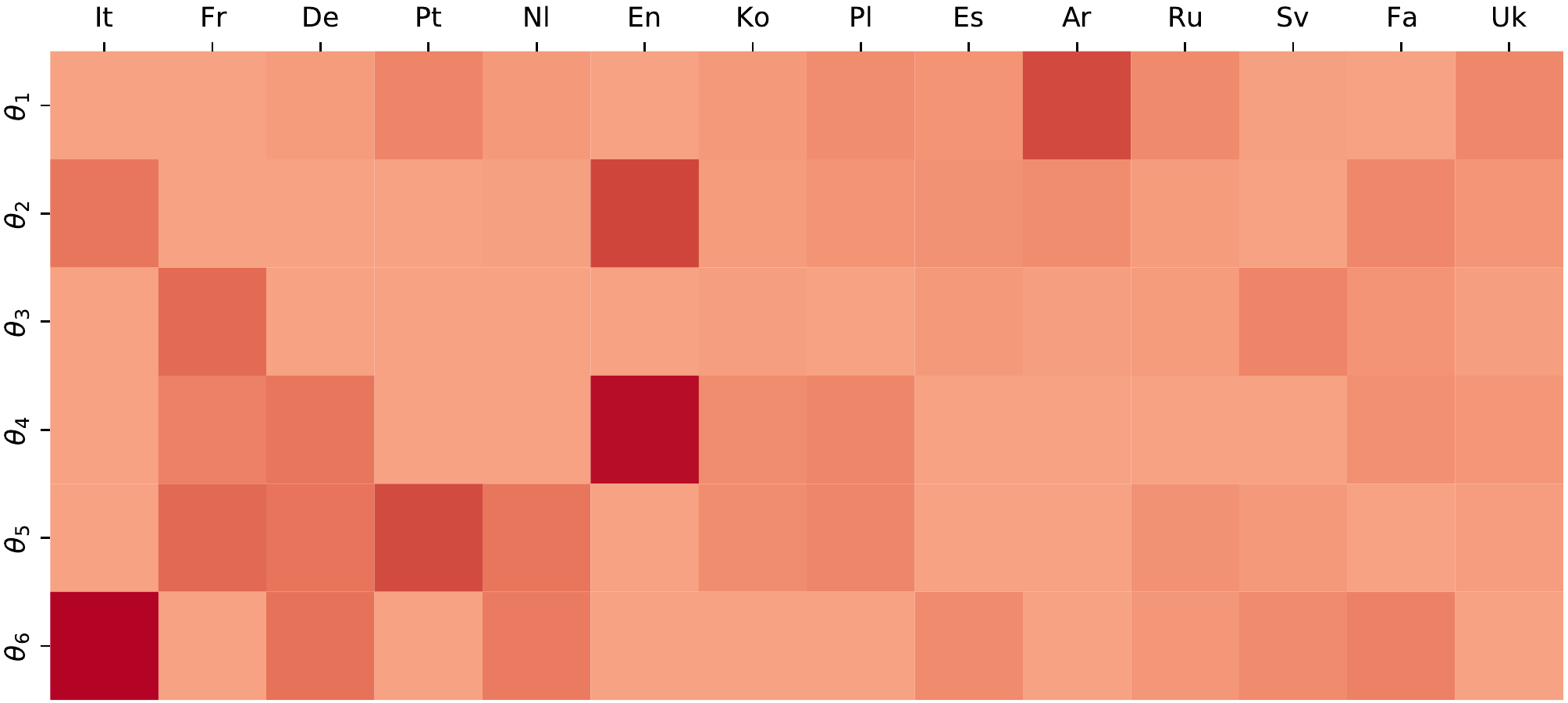}
\caption{The selection probability distributions of 6 sub-modules from LS on 14 languages. The sub-modules $\{\theta\}_{1}^{6}$ are numbered from 1 to 6. The languages are ordered from high-resource languages (left) to low-resource languages (right). The darker color means a higher selection probability to the corresponding sub-module and a lower probability to select a certain sub-module when the color is lighter.  }\label{fig8}
\end{figure}

\noindent\textbf{Selection Mechanism}
To observe how the selection mechanism affects our model performance, we also train one-to-one sub-modules of LS called \ourmethod{}14 without using the selection mechanism in the second training stage. Each independent sub-module corresponds to a language and each sentence is routed via a language prefix which represents the number of sub-module. We can visualize from Figure \ref{fig7} that increasing the number of parameters also improves obviously over \ourmethod{}-LS-LA. Nonetheless, the \ourmethod{}14 will suffer from the sharp increasing training time and inference time, and big space consumption when the number of languages is large enough. Instead of increasing parameters, our Language-specific Switcher can effectively ameliorate extraction quality with only slight extra parameters and less time consumption. Since similar languages tend to select the same sub-modules from our LS. The \ourmethod{} saves nearly 700M model capacity in our statistics and achieves better performance among most languages compared with \ourmethod{}14. It is obvious that \ourmethod{} is light and easy to transfer to other multi-field tasks.
\noindent\textbf{Selection Distribution}
Figure \ref{fig8} illustrates the heatmap of selection probability on 6 sub-modules from LS for each language. 
For each sub-module from top to bottom in Figure \ref{fig8}, we can visualize $\theta_1$ pays more attention to low-resource languages while $\theta_4$, $\theta_5$, and $\theta_6$ pay more attention to languages from the EURO family, which are mostly high-resource languages. The $\theta_2$ and $\theta_3$ seem to be more balanced on parameters sharing of languages except for 2 or 3 prominent languages.
We conclude that some sub-modules are mainly used to extract features from similar languages and others are used to assist the specific languages.

For each language from left to right, we can visualize that the selected sub-modules with higher probabilities are easy to distinguish in high-resource languages. In contrast, the selection probabilities across all sub-modules are relatively similar on low-resource languages in total.
We conclude that training data is rich enough to determine which way to route on high-resource languages and a more balanced selection decision is made on less training data.
It is learned from Figure \ref{fig8} that there are nearly 3 out of 6 prominently higher selection probabilities on the high-resource languages, and so do the low-resource languages with careful observation. It proves that only 3 sub-modules play the dominant role in refining the language-specific feature for each language.
To avoid interference from the other irrelevant sub-modules, we adopt a top-$K$ strategy to filter out $6-k$ sub-modules with lower selection probabilities in the evaluation stage. 
The top-$6$ strategy means selecting all sub-modules, which is the same as the training stage and the performance is relatively low (77.74) on average, while our \ourmethod{} achieves the best performance (77.87) when adopting the top-$3$ strategy. 
It demonstrates that filtering out the least important sub-modules is necessary to enhance the prediction quality, which also reduces the redundant parameters in the evaluation. The top-$1$ achieves the worst performance (77.60), which demonstrates the part of sub-modules are also helpful for the task. Therefore, the best performance is obtained when the $k$ value is balanced in all languages.
\noindent\textbf{Layer Number of Language-specific Switcher}
Table \ref{tab3} used to evaluate the effect of the layer number of LS. We divide the 6 sub-modules into 2 groups (each group has the same layer number) with different combinations of layer numbers to accommodate the scenarios, such as high- and low-resource language feature extraction. From Table \ref{tab3}, we can observe that the combination 1-2 achieves the best $F_1$ score on average. The combinations which are set to 1-1 and 4-4 also achieve better performance. With the increase or decrease of the layer number to a certain degree, the performances are almost the same, which maintains relatively low averaged $F_1$ scores. The full layer number combination 4-4 is an exception in the case, which demonstrates the performance still can be improved when the model capacity is large enough. According to the outcomes from Table \ref{tab3}, we conclude that the layer number of LS obviously impacts the results, with the best results attained when a balance is reached.

\begin{table}[h]
\begin{center}
\caption{The different layer numbers of sub-modules. Every 3 sub-modules in a group has the same layer numbers. Layer Num.01 and Layer Num.02 denote the layer number of the first group and second group respectively.}  \label{tab3}%
\resizebox{\linewidth}{!}{ 
\begin{tabular}{@{}cc|c|cccccccccccccc@{}}
\toprule
  Layer Num.01 & Layer Num.02 & AVG  & IT & FR & DE & PT & NL & EN & KO  & PL & ES & AR & RU & SV & FA & UK \\
\midrule
  1 & 1 & 77.4 & 81.3 & 69.1 & 72.1 & 73.4 & 80.4 & \textbf{63.1} & 81.9 & 81.3 & 63.4 & 91.1 & \textbf{85.4} & \textbf{83.7} & 77.8 & \textbf{80.0} \\
  \textbf{1} & \textbf{2} & \textbf{77.9} & \textbf{81.7} & 70.3 & \textbf{73.4}  & \textbf{74.3} & 81.1 & 62.3 & 82.7 & \textbf{81.6}  & 64.7 & \textbf{91.6} & 83.1  & \textbf{83.7} & \textbf{79.6} & \textbf{80.0} \\
  1 & 3 & 77.5 & 81.5 & 69.4 & 72.8 & 73.8 & 81.1 & 62.2 & 81.9 & 81.3 & 64.7 & 90.5 & 83.1 & \textbf{83.7} & \textbf{79.6} & \textbf{80.0} \\
  1 & 4 & 77.5 & 81.0 & 70.2 & 72.5 & 74.1 & 80.8 & 62.2 & 82.2 & 81.3 & 65.6 & 90.5 & 83.1 & \textbf{83.7} & 77.8 & \textbf{80.0} 
  \\
  2 & 2 & 77.8 & \textbf{81.7} & 70.1 & 73.1 & 74.1 & 81.0 & 62.3 & 81.9 & \textbf{81.6} & \textbf{66.1} & 90.5 & 83.1 & \textbf{83.7} & \textbf{79.6} & \textbf{80.0} \\
  2 & 3 & 77.4 & 81.1 & 70.2 & 72.7 & 74.1 & 80.9 & 62.1 & 82.5 & 81.3 & 65.2 & 90.5 & 81.5 & \textbf{83.7} & 77.8 & \textbf{80.0} 
  \\
  2 & 4 & 77.4 & 81.3 & 70.2 & 72.2 & 74.2 & 81.0 & 62.2 & 81.4 & 81.0 & 64.7 & 90.5 & 81.5 & \textbf{83.7} & \textbf{79.6} & \textbf{80.0} \\
  3 & 3 & 77.4 & 81.6 & \textbf{70.4} & 72.5 & 73.3 & 81.0 & 62.1 & 82.7 & 81.0 & 64.7 & 91.1 & 81.5 & \textbf{83.7} & 77.8 & \textbf{80.0} \\
  3 & 4 & 77.5 & 81.5 & 70.2 & 73.3 & 73.9 & \textbf{81.4} & 62.7 & \textbf{83.0} & 81.0 & 64.3 & 90.5 & 82.3 & \textbf{83.7} & 77.8 & \textbf{80.0} \\
  4 & 4 & 77.7 & 81.1 & 70.2 & 73.1 & 74.2 & 80.9 & 62.3 & 81.9 & \textbf{81.6} & 65.6 & 90.5 & 83.1 & \textbf{83.7} & \textbf{79.6} & \textbf{80.0} \\

\bottomrule
\end{tabular}
}
\end{center}
\end{table}

\section{Conclusion}
In this paper, we introduce a two-stage training method and a robust framework called \ourmethod{} for multilingual entity and relation extraction, which ameliorates the sentence representation quality and mitigates the language interference among multiple languages. Specifically, we first learn the generalities across all languages to obtain the unified language representation via the Language-universal Aggregator and then learn the specialties  of each language via the Language-specific Switcher. Experimental results demonstrate that our method significantly outperforms both monolingual and multilingual ERE baselines, which demonstrates that our framework can extract relational triples among various languages well. Moreover, our framework is also light and easy to transfer to other backbone models of multi-field tasks.

In the future, we will pay more attention to complex multilingual relational triple extraction, such as overlapping relational triples or multiple relational triples. Besides, we will also do further research on a better contextual representation among multiple languages. Although there is a long way to experience in multilingual entity and relation extraction tasks, it is important to investigate the valuable structured information in many other languages for the downstream NLP tasks.

\backmatter

\bmhead{Acknowledgments}
This work was supported in part by the National Natural Science Foundation of China (Grant Nos. 62276017, U1636211, 61672081), the 2022 Tencent Big Travel Rhino-Bird Special Research Program, and the Fund of the State Key Laboratory of Software Development Environment (Grant No. SKLSDE-2021ZX-18).





\bibliography{sn_bibliography}


\end{document}